\documentclass[letterpaper]{article} 
\usepackage{aaai23}  
\usepackage{times}  
\usepackage{helvet}  
\usepackage{courier}  
\usepackage[hyphens]{url}  
\usepackage{graphicx} 
\usepackage{natbib}  
\usepackage{caption} 
\usepackage{algorithm}
\usepackage{algorithmic}

\usepackage[utf8]{inputenc} 
\usepackage[T1]{fontenc}    
\usepackage{url}            
\usepackage{booktabs}       
\usepackage{amsfonts}       
\usepackage{nicefrac}       
\usepackage{microtype}      
\usepackage{xcolor}         

\usepackage{graphicx}       
\usepackage{amsmath}
\usepackage{amssymb}
\usepackage{multicol}
\usepackage{makecell}
\usepackage{multirow}
\usepackage{enumitem,kantlipsum}
\usepackage{wrapfig,lipsum,booktabs}

\usepackage{bbold}
\usepackage{mathtools}

\newcommand{\tabincell}[2]{\begin{tabular}{@{}#1@{}}#2\end{tabular}}

\DeclareMathOperator*{\argmin}{arg\,min}

\newcommand{\eg}{\textit{e.g.}~}

\newcommand{\ie}{\textit{i.e.}~}

\usepackage{newfloat}
\usepackage{listings}
\DeclareCaptionStyle{ruled}{labelfont=normalfont,labelsep=colon,strut=off} 
\floatstyle{ruled}
\newfloat{listing}{tb}{lst}{}
\floatname{listing}{Listing}
\title{Incremental-DETR: Incremental Few-Shot Object Detection via Self-Supervised Learning}
\author {
    Na Dong\textsuperscript{\rm 1,\rm 2}\thanks{Work fully done while first author is a visiting PhD student at the National University of Singapore.}, 
    Yongqiang Zhang\textsuperscript{\rm 2}, 
    Mingli Ding\textsuperscript{\rm 2}, 
    Gim Hee Lee\textsuperscript{\rm 1}
}
\affiliations {
    \textsuperscript{\rm 1} Department of Computer Science, National University of Singapore\\
    \textsuperscript{\rm 2} School of Instrument Science and Engineering, Harbin Institute of Technology\\
    { \{dongna1994, zhangyongqiang, dingml\}@hit.edu.cn,  gimhee.lee@comp.nus.edu.sg}
}

\usepackage{bibentry}

\begin{document}

\maketitle

\begin{abstract}
Incremental few-shot object detection aims at detecting novel classes without forgetting knowledge of the base classes with only a few labeled training data from the novel classes. Most related prior works are on incremental object detection that rely on the availability of abundant training samples per novel class that substantially limits the scalability to real-world setting where novel data can be scarce.
In this paper, we propose the Incremental-DETR that does incremental few-shot object detection via fine-tuning and self-supervised learning on the DETR object detector. To alleviate severe over-fitting with few novel class data, we first fine-tune the class-specific components of DETR with self-supervision from additional object proposals generated using Selective Search as pseudo labels. We further introduce an incremental few-shot fine-tuning strategy with knowledge distillation on the class-specific components of DETR to encourage the network in detecting novel classes without forgetting the base classes.    
Extensive experiments conducted on standard incremental object detection and incremental few-shot object detection settings show that our approach significantly outperforms state-of-the-art methods by a large margin.  Our source code is available at \url{https://github.com/dongnana777/Incremental-DETR}.
\end{abstract}

\section{Introduction}
\label{Introduction}
In the past decade, many impressive general object detectors have been developed due to the huge success of deep learning~\cite{girshick2013rich,uijlings2013selective,girshick2015fast,Lin2017Feature,redmon2015you,Liu2016SSD,lin2017focal,carion2020end, zhu2020deformable}. However, most deep learning-based object detectors can do detection on objects only from a fixed set of base classes that are seen during training.
The extension of the object detector to additional unseen novel classes without losing performance on the base classes requires further training on large amounts of training data from both novel and base classes.
A naive fine-tuning with training data only from the novel classes can lead to the notorious catastrophic forgetting problem, where knowledge of the base classes is quickly forgotten when the training data from the base classes are no longer available~\cite{mccloskey1989catastrophic, ratcliff1990connectionist}. 
Additionally, these deep learning-based detectors also suffer from the severe over-fitting problem when large amounts of annotated training data that are costly and tedious to obtain becomes scarce. 
In contrast, humans are much better than machines in continually learning novel concepts without forgetting previously learned knowledge despite the absence of previous examples and the availability of only a few novel examples. This gap between humans and machine learning algorithms fuels the interest in incremental few-shot object detection, which aims at continually extending the model to novel classes without forgetting the base classes with only a few samples per novel class.

In this paper, we focus on solving the problem of incremental few-shot object detection.
To this end, we propose the Incremental-DETR that does incremental few-shot object detection via fine-tuning and self-supervised learning on the recently proposed DETR object detector~\cite{zhu2020deformable}.  
We are inspired by the fine-tuning technique commonly used in few-shot object detectors~\cite{wang2020frustratingly,wu2020multi,sun2021fsce} based on the two-stage Faster R-CNN framework with class-agnostic feature extractor and Region Proposal Network (RPN). In the first stage, the whole network is trained on abundant base data. In the second stage, the class-agnostic feature extractor and RPN are frozen, and only the prediction heads are fine-tuned on a balanced subset that consists of both base and novel classes. However, many few-shot object detectors focus on detecting the novel classes but fail to preserve the performance of the base classes. In contrast, we consider the more challenging and practical incremental few-shot object detection which is capable of detecting both the novel and base classes. Furthermore, the data of the base classes are no longer accessible when the novel classes are introduced in incremental few-shot object detection. However, the training data of base classes are still accessible when training the novel model in few-shot object detection by which the novel model is easier to maintain the performance of the base classes.  

The state-of-the-art DETR object detector consists of a CNN backbone that extracts features from input images, a projection layer that reduces the channel dimension of features, an encoder-decoder transformer that transforms the features into features of a set of object queries, a 3-layer feed forward network that acts as the regression head, and a linear projection that acts as the classification head. 
We 
unfreeze different layers of DETR for fine-tuning, and empirically identify that the projection layer and classification head are class-specific and the CNN backbone, transformer and regression head of DETR are class-agnostic. The key part of our method is to separate the training of the class-agnostic and the class-specific components of DETR into two stages: 1) base model pre-training and self-supervised fine-tuning, and 2) incremental few-shot fine-tuning.  
Specifically, the whole network is pre-trained on abundant data from the base classes in the first part of the first stage. In the next part of the first stage, we propose a self-supervised learning method to fine-tune the class-specific projection layer and classification head together with the available abundant base class data.
In the second stage, the class-agnostic CNN backbone, transformer and regression head are kept frozen. We fine-tune the class-specific projection layer and classification head on a few examples of only the novel classes. Catastrophic forgetting is mitigated by identifying and freezing the class-agnostic components in both stages. 
Our fine-tuning of the class-specific components in both stages alleviates the over-fitting problem while giving the network ability to detect novel class objects.

Intuitively, humans can easily observe and extract additional meaningful object proposals in the background of the base class images from their prior knowledge. Based on this intuition, we leverage the selective search algorithm~\cite{uijlings2013selective} to generate additional object proposals that may exist in the background as pseudo ground truths to fine-tune the network in the first stage to detect the class-agnostic region proposals. 
Specifically, we do multi-task learning only on the class-specific projection layer and classification head, where the generate pseudo ground truths are used to self-supervise the model alongside the full-supervision from the abundant base class data. This helps the model to learn a more generalizable and transferable projection layer of DETR for incremental few-shot object detection and generalize to novel classes without forgetting the base classes.  
Furthermore, we propose a classification distillation loss and a masked feature distillation loss in addition to the existing loss of DETR for our incremental few-shot fine-tuning stage to impede catastrophic forgetting. Our \textbf{main contributions} are as follows:
\begin{itemize}
\item We propose the Incremental-DETR to tackle the challenging and relatively under-explored incremental few-shot object detection problem. 
\item We identify that the projection layer and the classification head of the DETR architecture are class-specific with empirical experimental studies. We thus adapt the fine-tuning strategy into our two-stage framework to avoid the catastrophic forgetting and over-fitting problems. 
\item We design a self-supervised method for the model to learn a more generalizable and transferable projection layer of DETR. Consequently, the model can efficiently adapt to the novel classes without forgetting the base classes in spite of novel class data scarcity.
\item Extensive experiments conducted on two standard object detection datasets (\ie MS COCO and PASCAL VOC) demonstrate the significant performance improvement of our approach over existing state-of-the-arts.
\end{itemize}


\begin{figure*}[!t]
\centering
\includegraphics[width=0.9\linewidth,height=7.2cm]{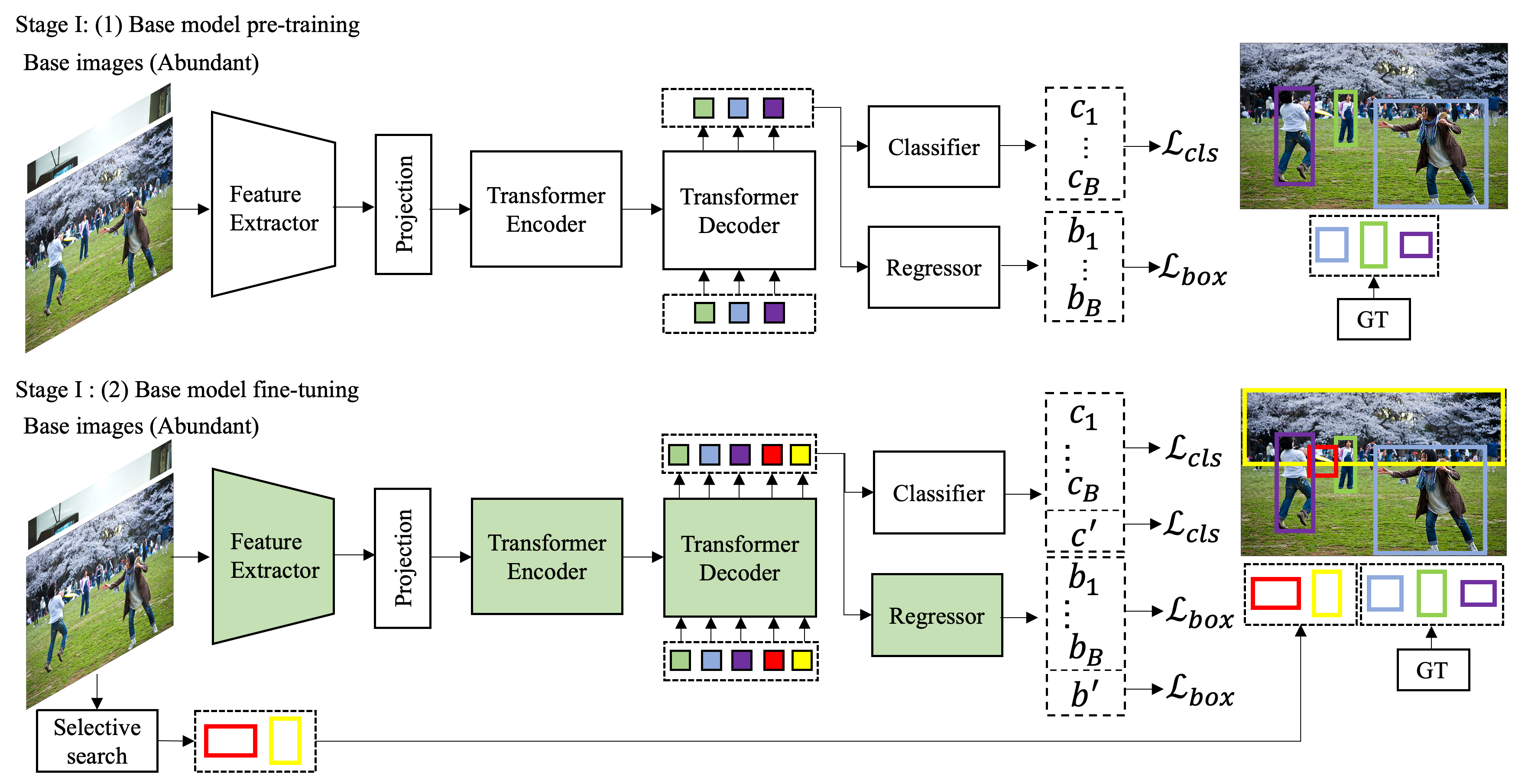}
\caption{Overview of our proposed base model training stage. Parameters of the modules shaded in green are frozen during training. Refer to the text for more details.
} 
\label{stage1}
\end{figure*}

\section{Related Works}
\label{Related Works}

\paragraph{Object Detection.}
Existing deep object detection models generally fall into two categories: 1) two-stage and 2) one-stage detectors. 
Two-stage detectors such as R-CNN~\cite{girshick2013rich} apply a deep neural network to extract features from proposals generated by selective search algorithm~\cite{uijlings2013selective}. Fast R-CNN~\cite{girshick2015fast}
utilizes a differentiable RoI Pooling to improve the speed and performance. Faster R-CNN~\cite{Ren2015Faster} introduces the Region Proposal Network (RPN) to generate proposals. FPN~\cite{Lin2017Feature} builds a top-down architecture with lateral connections to extract features across multiple layers.
In contrast, one-stage detectors such as YOLO~\cite{redmon2015you} directly perform object classification and bounding box regression on the features. SSD~\cite{Liu2016SSD} uses feature pyramid with different anchor sizes to cover the possible object scales. RetinaNet~\cite{lin2017focal} proposes the focal loss to mitigate the imbalanced positive and negative examples. Recently, another type of object detection methods~\cite{carion2020end, zhu2020deformable} beyond the one-stage and two-stage methods
have gained popularity. They directly supervise bounding box predictions end-to-end with Hungarian bipartite matching. 
However, these detectors require large amounts of training images per class and need to train the detectors over many training epochs and thus suffer from catastrophic forgetting of base classes and over-fitting in the context of incremental few-shot learning.
Therefore, it is imperative to extend the capability of the detectors to novel categories with no access to the original base training data with only a few samples. 


\paragraph{Few-Shot Object Detection.}
Several earlier works in few-shot object detection utilize the meta-learning strategy. MetaDet~\cite{wang2019meta},
MetaYOLO~\cite{kang2019few}, Meta R-CNN~\cite{wu2020meta} and Meta-DETR~\cite{zhang2021meta} use the meta-learner to generate a prototype per category from the support data and aggregate these prototypes with the query features by channel-wise multiplication. 
In contrast to the meta-learning based strategy, two-stage fine-tuning based methods show more potential in improving the performance of few-shot object detection.
For example, TFA~\cite{wang2020frustratingly} first trains Faster R-CNN on the base classes and then only fine-tunes the predictor heads. MPSR~\cite{wu2020multi} mitigates the scale scarcity in the few-shot datasets. FSCE~\cite{sun2021fsce} resorts to contrastive learning to learn discriminative object proposal representations.
Our proposed Incremental-DETR falls under the category of two-stage fine-tuning strategy, but differs from existing approaches that build upon Faster R-CNN for few-shot detection. Specifically, we consider the more challenging and practical incremental few-shot object detection by incorporating the two-stage fine-tuning strategy into the recently proposed DETR framework. 
Incremental few-shot object detection aims at learning a model of novel classes without forgetting the base classes with only a few samples per novel class. In contrast, many few-shot object detection works focus on detecting the novel classes but fail to preserve the performance of the base classes. Furthermore, the training data of the base classes are no longer accessible when the novel classes are introduced in the incremental few-shot object detection. However, the training data of base classes are still accessible when training the novel model in few-shot object detection by which the novel model is easier to maintain the performance of the base classes.  

\paragraph{Incremental Few-Shot Object Detection.}
Incremental few-shot object detection is first explored in ONCE~\cite{perez2020incremental}, which uses CenterNet~\cite{zhou2019objects} as a backbone to learn a class-agnostic feature extractor and a per-class code generator network for the novel classes. 
CenterNet~\cite{zhou2019objects} is well-suited for incremental few-shot object detection since a separate heatmap that makes independent detection by an activation thresholding is maintained for each individual object class. 
MTFA~\cite{ganea2021incremental} extends TFA~\cite{wang2020frustratingly} in a similar way as Mask R-CNN extends Faster R-CNN. MTFA is then adapted into incremental MTFA (iMTFA) by learning discriminative embeddings for object instances that are merged into class representatives. 
In this paper, we do incremental few-shot object detection on the DETR object detector.
However, DETR does not have a separate heatmap for each class and cannot be easily decoupled to independent detection for each class. Furthermore, DETR is a deeper network, and thus significantly more training data and computation time are needed for convergence. Since we only have access to a few samples of the  novel classes in the incremental few-shot setting and the data of base classes are no longer accessible, DETR is easier to catastrophically forget the base classes due to the absence of the base data and overfit to the novel classes due to the novel data scarcity. 


\begin{figure*}[!t]
\centering
\includegraphics[width=0.9\linewidth,height=5.2cm]{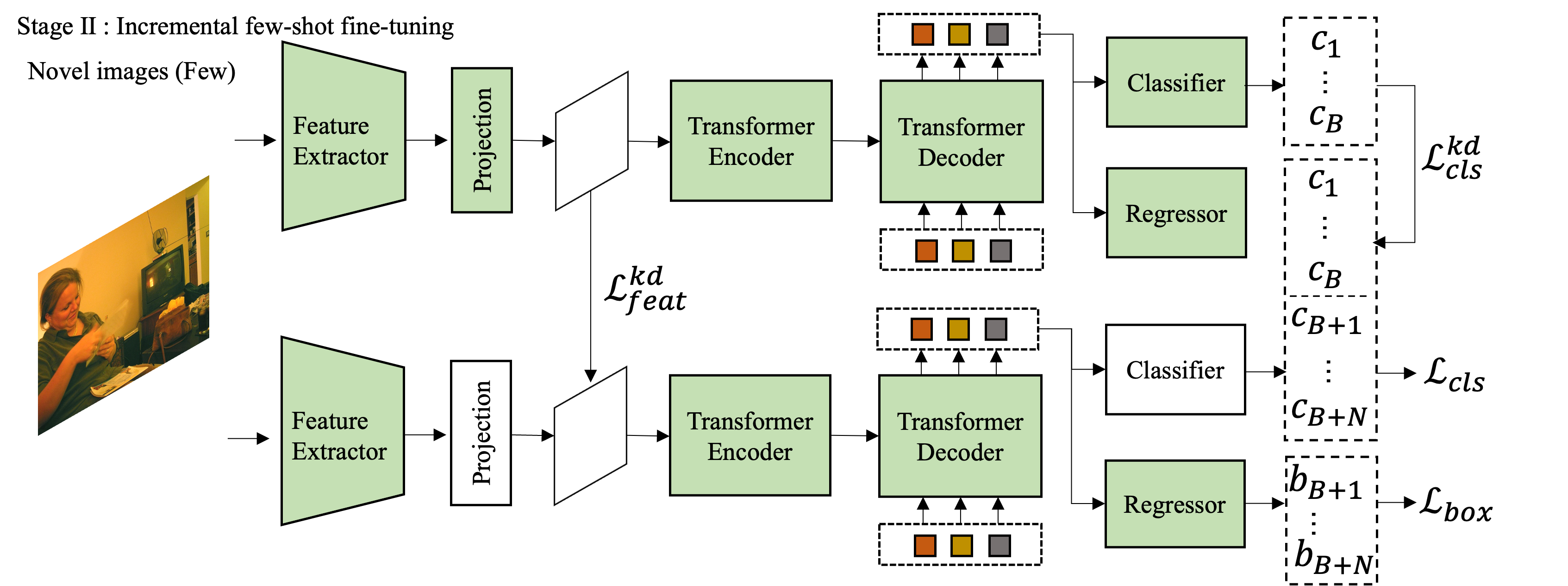}
\caption{Overview of our proposed incremental few-shot fine-tuning stage. 
Parameters of the modules shaded in green are frozen during training. Refer to the text for more details.
} 
\label{stage2}
\end{figure*}

\section{Problem Definition}
Let $\mathit{(x,y)} \in \mathcal{D}$ denotes a dataset $\mathcal{D}$ which contains images $\mathit{x}$ and their corresponding ground truth set of objects $\mathit{y}$. We further denote the training dataset of the base classes and the novel classes
as $\mathcal{D}_\text{base}$ and $\mathcal{D}_\text{novel}$, respectively.
Following the definition of class-incremental learning, we only have access to the novel class data $\mathcal{D}_\text{novel}$, where $y_\text{novel} \in \{ \mathit{C}_{B+1}, \dots, \mathit{C}_{B+N}\}$. The base class data $\mathcal{D}_\text{base}$, where $y_\text{base} \in \{\mathit{C_1}, \dots, \mathit{C_B}\}$ are no longer accessible when novel classes are introduced.
$\mathcal{D}_\text{base}$ and $\mathcal{D}_\text{novel}$ have no overlapped classes, \ie $ \{\mathit{C_1}, \dots, \mathit{C_B}\} \cap \{ \mathit{C}_{B+1}, \dots, \mathit{C}_{B+N}\} = \varnothing$. 
The training dataset $\mathcal{D}_\text{base}$ has abundant annotated instances of $ \{\mathit{C_1}, \dots, \mathit{C_B}\}$. In contrast, $\mathcal{D}_\text{novel}$ has very few annotated instances of  $\{ \mathit{C}_{B+1}, \dots, \mathit{C}_{B+N} \}$ and are often described as a $N$-way $K$-shot training set, where there are $N$ novel classes and each novel class has $K$ annotated object instances. 
The goal of incremental few-shot object detection is to continually learn novel classes $\{\mathit{C}_{B+1}, \dots, \mathit{C}_{B+N}\}$ from only a few training examples without forgetting knowledge of the base classes $\{\mathit{C_1}, \dots, \mathit{C_B}\}$.

\section{Our Methodology}
\label{Methodology}


\subsection{Base Model Training}
As shown in Figure~\ref{stage1}, we split the first stage of our incremental-DETR into base model pre-training and fine-tuning. 
In base model pre-training, we train the whole model only on the base classes $\{\mathit{C_1}, \dots, \mathit{C_B}\}$ with the same loss functions used in DETR.
Following DETR, we denote the set of $M$ predictions for base classes as $\hat{y} = \{ \hat{y}_i \}^M_{i=1} = \{  (\hat{c}_i, \hat{\mathbf{b}}_i)\}^M_{i=1}$, and the ground truth set of objects as $y = \{ y_i \}^M_{i=1} = \{(c_i, \mathbf{b}_i)\}^M_{i=1}$ padded with $\varnothing$ (no object). For each element $i$ of the ground truth set, $c_i$ is the target class label (which may be $\varnothing$ ) and $\mathbf{b}_i \in [0, 1]^4$ is a 4-vector that defines ground truth bounding box center coordinates, and its height and width relative to the image size. 
We adopt a pair-wise matching cost $\mathcal{L}_\text{match}(y_i, \hat{y}_{\sigma(i)})$ between the ground truth $y_i$ and a prediction $\hat{y}_{\sigma(i)}$ with index $\sigma(i)$ to search for a bipartite matching with the lowest cost:
\begin{equation}
\begin{array}{rll}
\begin{aligned}
\displaystyle 
\mathcal{\hat{\sigma}}= \argmin_{\sigma} \sum^M_{i=1} \mathcal{L}_\text{match}(y_i, \hat{y}_{\sigma(i)}).
\end{aligned}
\end{array}
\end{equation}
The matching cost takes into account both the class prediction $\hat{c}_{\sigma(i)}$, and the similarity of the predicted $\hat{\mathbf{b}}_{\sigma(i)}$ and ground truth $\mathbf{b}_i$ boxes. Specifically, it is defined as:

\begin{equation}
\begin{array}{rll}
\begin{aligned}
\displaystyle 
\mathcal{L}_\text{match}(y_i, \hat{y}_{\sigma(i)}) &= \mathbb{1}_{\{c_i \neq \varnothing \}} \mathcal{L}_\text{cls}(c_i, \hat{c}_{\sigma(i)}) \\ &+  \mathbb{1}_{\{c_i \neq \varnothing \}} \mathcal{L}_\text{box}(\mathbf{b}_i, \hat{\mathbf{b}}_{\sigma(i)}).
\end{aligned}
\end{array}
\end{equation}

Given the above definitions, the Hungarian loss \cite{kuhn1955hungarian} for all pairs matched in the previous step is defined as:
\begin{equation}
\begin{array}{rll}
\begin{aligned}
\displaystyle 
\mathcal{L}_\text{hg} (y, \hat{y}) = \sum_{i=1}^M [\mathcal{L}_\text{cls}(c_i, \hat{c}_{\hat{\sigma}(i)}) +  \mathbb{1}_{\{c_i \neq \varnothing \}} \mathcal{L}_\text{box}(\mathbf{b}_i, \hat{\mathbf{b}}_{\hat{\sigma}(i)})],
\end{aligned}
\end{array}
\end{equation}
where $\hat{\sigma}$ denotes the optimal assignment between predictions and targets. $\mathcal{L}_\text{cls}$ is the sigmoid focal loss \cite{lin2017focal}. $\mathcal{L}_\text{box}$ is a linear combination of $\ell_1$ loss and generalized IoU loss \cite{rezatofighi2019generalized} with the same weight hyperparameters as DETR. 

To make the parameters of class-specific components of DETR generalize well to the novel classes with a few samples in the incremental few-shot fine-tuning stage, we propose to fine-tune the class-specific components in a self-supervised way while keeping the class-agnostic components frozen in the base model fine-tuning of the first stage, where the parameters of the model is initialized from the pre-trained base model. 
The base model fine-tuning relies on making predictions on the other potential class-agnostic objects of base data along with ground truth objects of base classes. 
Inspired by R-CNN~\cite{girshick2014rich} and Fast R-CNN~\cite{girshick2015fast}, we use the selective search algorithm~\cite{uijlings2013selective} to generate a set of class-agnostic object proposals for each of the raw images. The selective search is a well-established and very effective unsupervised object proposal generation algorithm which uses color similarity, texture similarity, size of region and fit between regions to generate object proposals.
However, the number of object proposals is large and the ranking is not precise. To circumvent this problem, we use the object ranking from the selective search algorithm to prune away imprecise object proposals.  
Specifically, we select the top $O$ objects in the ranking list that are also not overlapping with the ground truth objects of the base classes as the pseudo ground truth object proposals. We represent the pseudo ground truth object proposals as $\mathbf{b}'$. To make predictions on these selected object proposals, we follow the prediction head of DETR to introduce a new class label $c'$ for all the selected object proposals as the pseudo ground truth label along with the labels of the base classes. Specifically, $c’$ depends on the number of categories of different sets. If there are n categories, we set $c’$ as $n+1$ along with the ground truth labels $(1, …, n)$.

Let us now denote the pseudo ground truth set of the selected object proposals as $y' = \{ y'_i \}^P_{i=1} = \{  (c'_i, \mathbf{b}'_i)\}^P_{i=1}$ padded with $\varnothing$ (no object) and the set of $P$ predictions as $\hat{y} = \{ \hat{y}_{i} \}^P_{i=1} = \{  (\hat{c}_i, \hat{\mathbf{b}}_{i})\}^P_{i=1}$. 
The same pair-wise matching cost $\mathcal{L}_\text{match}(y'_i, \hat{y}_{\sigma'(i)})$ between a pseudo ground truth $y'_i$ and a prediction $\hat{y}_{\sigma'(i)}$ with index $\sigma'(i)$ is also adopted to search for a bipartite matching with the lowest cost.
Finally, the Hungarian loss for all the matched pairs is defined as:
\begin{equation}
\begin{array}{rll}
\begin{aligned}
\displaystyle 
\mathcal{L}_\text{hg} (y', \hat{y}) = \sum_{i=1}^P [\mathcal{L}_\text{cls}(c'_i, \hat{c}_{\hat{\sigma}'(i)}) +  \mathbb{1}_{\{c'_i \neq \varnothing \}} \mathcal{L}_\text{box}(\mathbf{b}'_i, \hat{\mathbf{b}}_{\hat{\sigma}'(i)})],
\end{aligned}
\end{array}
\end{equation}
where $\hat{\sigma}'$ denotes the optimal assignment between predictions and the pseudo ground truths. $\mathcal{L}_\text{cls}$ is the sigmoid focal loss which does binary classification between the selected object proposals and the background.
$\mathcal{L}_\text{box}$ is a linear combination of $\ell_1$ loss and generalized IoU loss with the same weight hyperparameters as DETR. 

The overall loss $\mathcal{L}_\text{total}^\text{base}$ to fine-tune the base model on the abundant base data $\mathcal{D}_\text{base}$ is given by:
\begin{equation}
\begin{array}{rll}
\begin{aligned}
\displaystyle 
\mathcal{L}_\text{total}^\text{base} = \mathcal{L}_\text{hg} (y, \hat{y}) + \lambda'\mathcal{L}_\text{hg} (y', \hat{y}),
\end{aligned}
\end{array}
\end{equation}
where $\lambda'$ is the hyperparameter to balance the loss terms. 

\subsection{Incremental Few-Shot Fine-Tuning}
As shown in Figure~\ref{stage2}, we first initialize the parameters of the novel model from the fine-tuned base model in the first stage. We then propose to fine-tune the class-specific projection layer and classification head with a few samples of the novel classes while keeping the class-agnostic components frozen. However, the constant updating of the projection layer and classification head in the process of novel classes learning can aggravate catastrophic forgetting of the base classes. Therefore, we propose to use knowledge distillation to mitigate catastrophic forgetting. Specifically, the base model is utilized to prevent the projection layer output features of the novel model from deviating too much from the projection layer output features of the base model. However, a direct knowledge distillation on the full features causes conflicts and thus hurts the performance of the novel classes.
We thus use the ground truth bounding boxes of the novel classes as a binary mask $\mathit{mask}^\text{novel}$  (1 within ground truth boxes, otherwise 0) to prevent negative influence on the novel class learning from features of the base model. The distillation loss with the mask on the features is written as:
\begin{equation}
\small
    \mathcal{L}_\text{feat}^\text{kd} =  \frac{1}{2N^\text{novel}} \sum_{i=1}^ w \sum_{j=1}^ h \sum_{k =1} ^c (1 - \mathit{mask}_{ij}^\text{novel}) \Big\|  \mathit{f}^\text{novel}_{ijk} - \mathit{f}^\text{base}_{ijk} \Big\| ^2,
\end{equation}
where  $N^\text{novel} = \sum_{i=1}^ w \sum_{j=1}^ h (1 - \mathit{mask}_{ij}^\text{novel})$. 
$\mathit{f^\text{base}}$ and $\mathit{f^\text{novel}}$  denote the features of the base and novel models, respectively. $\mathit{w}$, $\mathit{h}$ and $\mathit{c}$ are the width, height and channels of the feature.

For the knowledge distillation on the classification head of DETR, we first select the prediction outputs from the $M$ prediction outputs of the base model as pseudo ground truths of the base classes. 
Specifically, for an input novel image, we consider a prediction output of the base model as a pseudo ground truth of the base classes when its class probability greater than 0.5 and its bounding box has no overlap with ground truth bounding boxes of the novel classes.  
We then adopt a pair-wise matching cost to find for the bipartite matching between the pseudo ground truths and the predictions of the novel model. Subsequently, the classification outputs of the base and novel models are compared in the distillation loss function given by:
\begin{equation}
\begin{array}{rll}
\begin{aligned}
\displaystyle 
\mathcal{L}_\text{cls}^\text{kd} &= 
 \mathcal{L}_\text{kl\_div} ( \log (\mathit{q^\text{novel}}), \mathit{q^\text{base}}),  \\
\end{aligned}
\end{array}
\end{equation}
where we follow \cite{hinton2015distilling} in the definition of the KL-divergence loss $\mathcal{L}_\text{kl\_div}$ between the class probabilities of the novel and base models. 
$\mathit{q}$ denotes the class probability.

The Hungarian loss $\mathcal{L}_\text{hg}$ is also applied to the ground truth set $y$ and the predictions $\hat{y}$ of the novel data $\mathcal{D}_\text{novel}$. The overall loss $\mathcal{L}_\text{total}^\text{novel}$ to train the novel model on the novel data $\mathcal{D}_\text{novel}$ is given by:
\begin{equation}
\begin{array}{rll}
\begin{aligned}
\displaystyle 
\mathcal{L}_\text{total}^\text{novel} = \mathcal{L}_\text{hg} (y, \hat{y}) + \lambda_\text{feat}\mathcal{L}_\text{feat}^\text{kd} + \lambda_\text{cls}\mathcal{L}_\text{cls}^\text{kd},
\end{aligned}
\end{array}
\end{equation}
where $\lambda_\text{feat}$ and $\lambda_\text{cls}$ are hyperparameters to balance the loss terms. 

\section{Experiments}
\label{Experiments}

\subsection{Experimental Setup}
\label{sec:ExperimentalSetup}
\subsubsection{Incremental object detection.}
We first evaluate the performance of our proposed method on the incremental setting as studied in~\cite{shmelkov2017incremental,kj2021incremental}.
We conduct the evaluation on the popular object detection benchmark MS COCO 2017~\cite{lin2014microsoft} which covers 80 object classes. The \textit{train} set of COCO serves as training data and the \textit{val} set serves as testing data. Standard evaluation metrics for COCO are adopted.
Following~\cite{shmelkov2017incremental, kj2021incremental}, we report the results on the incremental learning setting of adding a group of novel classes (40+40). 
Furthermore, results on the setting of adding one novel class (40+1) are also reported to prove the effectiveness of our method.

\subsubsection{Incremental few-shot object detection.} We follow the data setups of prior works on incremental few-shot object detection, \eg~\cite{perez2020incremental}. Specifically, we conduct the experimental evaluations on two widely used object detection benchmarks: MS COCO 2017 and PASCAL VOC 2007~\cite{everingham2010pascal}. The \textit{train} set of COCO serves as training data and the \textit{val} set serves as testing data. The \textit{trainval} set of VOC serves as training data and the \textit{test} set serves as testing data. Standard evaluation metrics for COCO are adopted.
COCO contains objects from 80 different classes including 20 classes that intersect with VOC. We adopt the 20 shared classes as novel classes and the remaining 60 classes as base classes. Following~\cite{perez2020incremental}, there are two dataset splits: same-dataset on COCO and cross-dataset with 60 COCO classes as base classes.
For the same-dataset evaluation on COCO, the model is pre-trained and fine-tuned on the COCO base training data. The novel classes learning is then conducted on COCO novel training data and evaluated on COCO testing data.
We use the same setup as above for the cross-dataset evaluation on COCO to VOC, where the model is pre-trained and fine-tuned on the COCO base training data. However, the incremental few-shot fine-tuning stage is conducted on the VOC training data and evaluated on the VOC testing data. We evaluate incremental few-shot object detection with $1, 5, 10$-shot per novel class and the base training data is no longer accessible during novel classes learning following the definition of class-incremental learning. 

\begin{table*}[!t]
\small
\centering
\resizebox{0.9\linewidth}{!}{
\begin{tabular}{p{1cm}<{\centering}|p{5.5cm}<{\centering}|p{1cm}<{\centering}p{1cm}<{\centering}|p{1cm}<{\centering}p{1cm}<{\centering}|p{1cm}<{\centering}p{1cm}<{\centering}}
\hline
\multirow{2}{*}{Shot}  & \multirow{2}{*}{Method} & \multicolumn{2}{c|}{Base}      & \multicolumn{2}{c|}{Novel}     & \multicolumn{2}{c}{All}       \\ \cline{3-8} 
                      &                       &                          \multicolumn{1}{p{2cm}<{\centering}|}{AP(\%)} & AP50(\%) & \multicolumn{1}{p{2cm}<{\centering}|}{AP(\%)} & AP50(\%) & \multicolumn{1}{p{2cm}<{\centering}|}{AP(\%)} & AP50(\%) \\ \hline
                       
\multirow{2}{*}{Base}                    
                  & iMTFA~\cite{ganea2021incremental}                       & \multicolumn{1}{c}{38.2}   &58.0      & \multicolumn{1}{c}{-}   &-      & \multicolumn{1}{c}{-} &- \\
             & \cite{zhu2020deformable}                         & \multicolumn{1}{c}{36.9}   &56.7      & \multicolumn{1}{c}{-}   &-      & \multicolumn{1}{c}{-} &- \\
             \hline

\multirow{3}{*}{1}      
      &ONCE~\cite{perez2020incremental}                         & \multicolumn{1}{c}{17.9}   & -     & \multicolumn{1}{c}{0.7}   &-     & \multicolumn{1}{c}{13.6}   & -     \\ 
          &iMTFA~\cite{ganea2021incremental}                       & \multicolumn{1}{c}{27.8}   &40.1     & \multicolumn{1}{c}{\textbf{3.2}}   &\textbf{5.9}      & \multicolumn{1}{c}{21.7}   &31.6      \\ 
                  &  Ours                        & \multicolumn{1}{c}{\textbf{29.4}}   & \textbf{47.1}     & \multicolumn{1}{c}{1.9}   &2.7      & \multicolumn{1}{c}{\textbf{22.5}}   & \textbf{36.0}     \\ \hline
   
\multirow{3}{*}{5}         
      &ONCE~\cite{perez2020incremental}                        & \multicolumn{1}{c}{17.9}   & -     & \multicolumn{1}{c}{1.0}   &-     & \multicolumn{1}{c}{13.7}   &-      \\ 
          &iMTFA~\cite{perez2020incremental}                        & \multicolumn{1}{c}{24.1}   & 33.7    & \multicolumn{1}{c}{6.1}   & 11.2     & \multicolumn{1}{c}{19.6}   & 28.1     \\ 
          &  Ours                        & \multicolumn{1}{c}{\textbf{30.5}}   &\textbf{48.4}      & \multicolumn{1}{c}{\textbf{8.3}}   & \textbf{13.3}     & \multicolumn{1}{c}{\textbf{24.9}}   &\textbf{39.6}      \\ \hline
                       
\multirow{3}{*}{10}       
        &ONCE~\cite{perez2020incremental}                         & \multicolumn{1}{c}{17.9}   & -     & \multicolumn{1}{c}{1.2}   &-     & \multicolumn{1}{c}{13.7}   & -     \\ 
                 &iMTFA~\cite{ganea2021incremental}                        & \multicolumn{1}{c}{23.4}   & 32.4     & \multicolumn{1}{c}{7.0}   & 12.7     & \multicolumn{1}{c}{19.3}   & 27.5     \\ 
                  &  Ours                        & \multicolumn{1}{c}{\textbf{27.3}}   & \textbf{44.0}     & \multicolumn{1}{c}{\textbf{14.4}}   &  \textbf{22.4}    & \multicolumn{1}{c}{\textbf{24.1}}   & \textbf{38.6}     \\ \hline

\end{tabular}}
\caption{\small Results of incremental few-shot object detection on COCO $val$ set. }
\label{fsod+ifsod on same dataset}
\end{table*}

\subsection{Implementation Details}
\label{sec:ImplementDetails}
We use ResNet-50 as the feature extractor. The network architectures and hyperparameters of transformer encoder and decoder remain the same as Deformable DETR~\cite{zhu2020deformable}.
$\lambda'$, $\lambda_\text{feat}$ and $\lambda_\text{cls}$ are set to 1, 0.1 and 2, respectively.
We report the results of our proposed method with single-scale feature. 
The training is carried out on 8 RTX 6000 GPUs with a batch size of 2 per GPU.
In the base model pre-training stage, we train our model using the AdamW optimizer and an initial learning rate of $2 \times 10^{-4}$ and a weight decay of $1 \times 10^{-4}$. We train our model for 50 epochs and the learning rate is decayed at 40$^\textup{{th}}$ epoch by a factor of $0.1$. 
In the base model fine-tuning stage, the model is initialized from pre-trained base model. The parameters of the projection layer and classification head are then fine-tuned while keeping the other parameters frozen. We fine-tune the model for 1 epoch with a learning rate of $2 \times 10^{-4}$. 
We apply the same setting as the base model pre-training stage in the incremental few-shot fine-tuning stage.

\subsection{Incremental Object Detection}

\subsubsection{Addition of one class.}
Table~\ref{40+1} shows the results on the incremental learning setting of one novel class addition.
We take the first 40 classes from MS COCO as the base classes, and the next 41$^\text{th}$ class is used as the novel class. 
We report the average precision (AP) and the average precision with 0.5 IoU threshold (AP50) of the base, novel and all classes, respectively. 
"1-41" and "1-40" are baselines trained without using incremental learning.  
"41 (fine-tune)" and "41 (scratch)" are the model trained with and without initialization from the pre-trained base model, respectively. The results of "41 (scratch)" are significantly lower than "41 (fine-tune)" due to the scarcity of the novel data.
Our approach achieves high results with AP and AP50 of $41.9\% \mid 64.1\%, 29.2\%\mid50.3\%, 41.6\%\mid 63.8\%$ for the base, novel and all classes, respectively.
Our results are almost on par with the results from baseline training on the novel class "41" with fine-tuning (Ours: $29.2\% \mid 50.3\% $ vs. "41 (fine-tune)": $30.7\% \mid 50.3\%$). Furthermore, our method also achieves results that are close to the baseline training on base classes "1-40" (Ours: $41.9\% \mid 64.1\%$ vs. "1-40": $44.8\% \mid 67.8\%$).
 These results support the effectiveness of our incremental learning framework in preserving base class knowledge and the learning of novel class knowledge.

\begin{table}[!htbp]
\large
\centering
\resizebox{\linewidth}{!}{
\begin{tabular}{p{1.3cm}<{\centering}|p{2.3cm}<{\centering}|p{1cm}<{\centering}p{1.1cm}<{\centering}|p{1cm}<{\centering}p{1.1cm}<{\centering}|p{1cm}<{\centering}p{1.1cm}<{\centering}}
\hline
\multirow{2}{*}{Class} & \multirow{2}{*}{Method} & \multicolumn{2}{c|}{Base}      & \multicolumn{2}{c|}{Novel}     & \multicolumn{2}{c}{All}       \\ \cline{3-8} 
                       &                         & \multicolumn{1}{c|}{AP(\%)} & AP50(\%) & \multicolumn{1}{c|}{AP(\%)} & AP50(\%) & \multicolumn{1}{c|}{AP(\%)} & AP50(\%) \\ \hline
                       
         1-41              &\cite{zhu2020deformable}                       & \multicolumn{1}{c}{45.6}   &68.6      & \multicolumn{1}{c}{31.3}   & 55.7     & \multicolumn{1}{c}{45.3}   &68.3     \\ 
          1-40              &\cite{zhu2020deformable}                       & \multicolumn{1}{c}{44.8}   & 67.8     & \multicolumn{1}{c}{-}   &-      & \multicolumn{1}{c}{-} &- \\
         \tabincell{c}{41 \\ (fine-tune) }            &\cite{zhu2020deformable}                       & \multicolumn{1}{c}{15.6}   &23.9      & \multicolumn{1}{c}{30.7}   &50.3      & \multicolumn{1}{c}{16.0} &24.6 \\    
         \tabincell{c}{ 41 \\ (scratch) }           &\cite{zhu2020deformable}                       & \multicolumn{1}{c}{-}   &-      & \multicolumn{1}{c}{16.7}   & 32.4     & \multicolumn{1}{c}{-} &- \\
          \hline
\tabincell{c}{(1-40) \\ + (41)}    
                       & Ours                        & \multicolumn{1}{c}{41.9}   &64.1      & \multicolumn{1}{c}{29.2}   &50.3      & \multicolumn{1}{c}{41.6}   & 63.8     \\ \hline
\end{tabular}}
\caption{\small Results of "40+1" on COCO $val$ set. "1-40" is the base classes, and "41" is the novel class.} 
\label{40+1}
\end{table}

\subsubsection{Addition of a group of classes.}
Table~\ref{40+40} shows the results on the first 40 classes from MS COCO as the base classes, and the next 40 classes as the additional group of novel classes.
"1-80" and "1-40" are baselines trained without using incremental learning. 
"41-80 (fine-tune)" and "41-80 (scratch)" are the model trained with and without initialization from the pre-trained base model, respectively.
The performance of "41-80 (scratch)" is better than "41-80 (fine-tune)" due to the abundance of novel class data. 
Our method achieves results with AP and AP50 that are almost on par with the baseline training on base classes "1-40" without using incremental learning (Ours: $44.0\% \mid 66.1\%$ vs. "1-40": $44.8\% \mid 67.8\%$). 
Additionally, our method significantly outperforms~\cite{kj2021incremental} on all classes (Ours: $37.3\% \mid 56.6\%$ vs. ~\cite{kj2021incremental}: $23.8\% \mid 40.5\%$). Our method also outperforms~\cite{shmelkov2017incremental} under the incremental setting (Ours: $37.3\% \mid 56.6\%$ vs. ~\cite{shmelkov2017incremental}: $21.3\% \mid 37.4\%$).
These results show the effectiveness of our approach when a group of novel classes is added. 

\begin{table}[!htbp]
\large
\centering
\resizebox{\linewidth}{!}{
\begin{tabular}{p{1.5cm}<{\centering}|p{2.5cm}<{\centering}|p{1cm}<{\centering}p{1.1cm}<{\centering}|p{1cm}<{\centering}p{1.1cm}<{\centering}|p{1cm}<{\centering}p{1.1cm}<{\centering}}
\hline
\multirow{2}{*}{Class} & \multirow{2}{*}{Method} & \multicolumn{2}{c|}{Base}      & \multicolumn{2}{c|}{Novel}     & \multicolumn{2}{c}{All}       \\ \cline{3-8} 
                       &                         & \multicolumn{1}{c|}{AP(\%)} & AP50(\%) & \multicolumn{1}{c|}{AP(\%)} & AP50(\%) & \multicolumn{1}{c|}{AP(\%)} & AP50(\%) \\ \hline
     1-80                  &\cite{zhu2020deformable}                       & \multicolumn{1}{c}{46.8}   &69.4      & \multicolumn{1}{c}{36.3}   & 54.7     & \multicolumn{1}{c}{41.4}   & 61.8     \\ 
     1-40                  &\cite{zhu2020deformable}                       & \multicolumn{1}{c}{44.8}   & 67.8     & \multicolumn{1}{c}{-}   &-      & \multicolumn{1}{c}{-} &- \\
     \tabincell{c}{41-80 \\ (fine-tune)}               &\cite{zhu2020deformable}                       & \multicolumn{1}{c}{0}   &0      & \multicolumn{1}{c}{33.0}   &49.6     & \multicolumn{1}{c}{16.5} &24.8 \\
     \tabincell{c}{ 41-80 \\ (scratch) }             &\cite{zhu2020deformable}                       & \multicolumn{1}{c}{-}   &-      & \multicolumn{1}{c}{35.0}   &52.6     & \multicolumn{1}{c}{-} &- \\
     \hline
                        
\multirow{3}{*}{ \tabincell{c}{(1-40) \\ + (41-80)} }      &    \cite{shmelkov2017incremental}                       & \multicolumn{1}{c}{-}   &-      & \multicolumn{1}{c}{-}   &-      & \multicolumn{1}{c}{21.3}   & 37.4     \\  
                       &  \cite{kj2021incremental}                        & \multicolumn{1}{c}{-}   &-      & \multicolumn{1}{c}{-}   &-      & \multicolumn{1}{c}{23.8}   &40.5      \\ 
                       & Ours                        & \multicolumn{1}{c}{44.0}   &66.1      & \multicolumn{1}{c}{30.6}   &47.1      & \multicolumn{1}{c}{\textbf{37.3}}   &\textbf{56.6}      \\ \hline
\end{tabular}}
\caption{\small Results of "40+40" on COCO $val$ set. "1-40" is the base classes, and "41-80" is the novel classes.} 
\label{40+40}
\end{table}

\subsection{Incremental Few-Shot Object Detection}

\begin{table*}[!t]
\small
\centering
\resizebox{0.9\linewidth}{!}{
\begin{tabular}{c|c|cc|cc|cc|cc|cc}
\hline
\multirow{2}{*}{Row ID} &\multirow{2}{*}{\tabincell{c}{Two-stage \\ fine-tuning}} 
&\multicolumn{2}{c|}{Self-supervised learning} 
&\multicolumn{2}{c|}{Knowledge distillation} 
& \multicolumn{2}{c|}{Base}      & \multicolumn{2}{c|}{Novel}     & \multicolumn{2}{c}{All}       \\ \cline{3-12} 

                       & &\multicolumn{1}{p{1.5cm}<{\centering}|}{$\mathcal{L}_\text{box}$} & $\mathcal{L}_\text{cls}$
                       &\multicolumn{1}{p{1.5cm}<{\centering}|}{$\mathcal{L}_\text{feat}^\text{kd}$} & $\mathcal{L}_\text{cls}^\text{kd}$ 
                       & \multicolumn{1}{c|}{AP(\%)} & AP50 (\%)
                       & \multicolumn{1}{c|}{AP(\%)} & AP50(\%) 
                       & \multicolumn{1}{c|}{AP(\%)} & AP50(\%) \\ \hline
                        1 &   &\multicolumn{1}{c}{}   &                        &  \multicolumn{1}{c}{}   &                                                  
                         & \multicolumn{1}{c}{0.1}   & 0.2     & \multicolumn{1}{c}{1.4}   &2.5      & \multicolumn{1}{c}{0.4}   & 0.8     \\

                 2     &\checkmark                        & \multicolumn{1}{c}{}   &                          &  \multicolumn{1}{c}{}   &                                            
                      & \multicolumn{1}{c}{19.7}   & 32.5     & \multicolumn{1}{c}{5.2}   &8.2      & \multicolumn{1}{c}{16.1}   & 26.4     \\ 
                    3 &\checkmark     & \multicolumn{1}{c}{\checkmark }                   &  \checkmark                        &  \multicolumn{1}{c}{}   &                                                    
                    & \multicolumn{1}{c}{16.3}   &27.1      & \multicolumn{1}{c}{8.0}   & 12.9     & \multicolumn{1}{c}{14.2}   & 23.5     \\ 
                       4 &\checkmark  &  \multicolumn{1}{c}{\checkmark}   &  \checkmark                                                              &   \multicolumn{1}{c}{\checkmark}   &                                                            & \multicolumn{1}{c}{23.8}   & 38.0     & \multicolumn{1}{c}{\textbf{8.3}}   &13.2      & \multicolumn{1}{c}{19.9}   & 31.8    \\ 
                        5 &\checkmark  &  \multicolumn{1}{c}{\checkmark}   &  \checkmark                                                              &   \multicolumn{1}{c}{}   &      \checkmark                                                      & \multicolumn{1}{c}{26.1}   & 43.0     & \multicolumn{1}{c}{8.0}   &12.7      & \multicolumn{1}{c}{21.6}   & 35.4    \\ 
                        6 &\checkmark  & \multicolumn{1}{c}{}   &           &  \multicolumn{1}{c}{\checkmark}   &  \checkmark                                                   & \multicolumn{1}{c}{\textbf{30.7}}   &\textbf{49.0}     & \multicolumn{1}{c}{5.1}   &8.5      & \multicolumn{1}{c}{24.3}   &38.9     \\ 
                        7 &\checkmark  &  \multicolumn{1}{c}{\checkmark}   &              &  \multicolumn{1}{c}{\checkmark}   &  \checkmark                                                   & \multicolumn{1}{c}{30.3}   &48.2      & \multicolumn{1}{c}{7.5}   & 12.2     & \multicolumn{1}{c}{24.6}   & 39.2     \\ 
                  
                       8 &\checkmark  &  \multicolumn{1}{c}{\checkmark}   &  \checkmark                         & \multicolumn{1}{c}{\checkmark}   &  \checkmark                                                     & \multicolumn{1}{c}{30.5}   &48.4      & \multicolumn{1}{c}{\textbf{8.3}}   &\textbf{13.3}      & \multicolumn{1}{c}{\textbf{24.9}}   & \textbf{39.6}     \\ 
                      
                        \hline
\end{tabular}}
\caption{\small Ablation studies with 5-shot per novel class on COCO $val$ set. } 
\label{ablation}
\end{table*}

\begin{table*}[!t]
\small
\centering
\resizebox{0.9\linewidth}{!}{
\begin{tabular}{p{1cm}<{\centering}|p{1.5cm}<{\centering}p{1.5cm}<{\centering}|p{1.5cm}<{\centering}p{1.5cm}<{\centering}|p{1cm}<{\centering}p{1cm}<{\centering}|p{1cm}<{\centering}p{1cm}<{\centering}|p{1cm}<{\centering}p{1cm}<{\centering}}
\hline
\multirow{2}{*}{Row ID} &\multicolumn{2}{c|}{\tabincell{c}{ Base model pre-training}} &\multicolumn{2}{c|}{\tabincell{c}{ Incremental few-shot fine-tuning}}

& \multicolumn{2}{c|}{Base}      & \multicolumn{2}{c|}{Novel}     & \multicolumn{2}{c}{All}       \\ \cline{2-11} 

                       & \multicolumn{1}{c|}{Two-step} & One-step 
                       & \multicolumn{1}{c|}{Frozen } & Unfrozen
                       & \multicolumn{1}{c|}{AP(\%)} & AP50(\%)
                       & \multicolumn{1}{c|}{AP(\%)} & AP50(\%) 
                       & \multicolumn{1}{c|}{AP(\%)} & AP50(\%) \\ \hline

                        1  &           &  \checkmark   &    \checkmark       &                                     
                                & \multicolumn{1}{c}{28.3}   & 45.2     & \multicolumn{1}{c}{7.1}   &11.3      & \multicolumn{1}{c}{23.0}   &36.7     \\             
                       
                          2 &\checkmark           &       &          &   \checkmark                                 & \multicolumn{1}{c}{4.3}   &9.8     & \multicolumn{1}{c}{3.5}   &6.2      & \multicolumn{1}{c}{4.1}   & 8.9    \\ 
                        3 &\checkmark           &       & \checkmark          &                                   & \multicolumn{1}{c}{\textbf{30.5}}   &\textbf{48.4}      & \multicolumn{1}{c}{\textbf{8.3}}   &\textbf{13.3}      & \multicolumn{1}{c}{\textbf{24.9}}   & \textbf{39.6}     \\ 
                         
                        \hline
\end{tabular}}
\caption{\small Ablation studies with 5-shot per novel class on COCO $val$ set. } 
\label{ablation2}
\end{table*}

\subsubsection{Same-dataset.}
In this experiment, we show the effectiveness of our method in learning to detect novel classes without forgetting base classes with only a few novel samples. 
The first two rows of Table~\ref{fsod+ifsod on same dataset} show the results of 
iMTFA~\cite{ganea2021incremental} and Deformable DETR~\cite{zhu2020deformable} trained on the base classes without using incremental few-shot learning. 
Our method achieves the best performance on the AP and AP50 for the base and all classes. Our method significantly outperforms iMTFA and ONCE~\cite{perez2020incremental} on the base, novel and all classes, except for 1-shot setting of the novel classes where all methods perform poorly due to the extreme data scarcity. Particularly, our method significantly outperforms iMTFA on the AP and AP50 for all classes, \ie iMTFA: ($21.7\% \mid 31.6\%, 19.6\% \mid 28.1\%, 19.3\% \mid 27.5\%$) vs Ours: ($22.5\% \mid 36.0\%, 24.9\% \mid 39.6\%, 24.1\% \mid 38.6\%$) under the 1, 5 and 10-shot settings, respectively.
These results show the superiority of our method for incremental few-shot object detection.

\subsubsection{Cross-dataset.}
We report the performance of incremental few-shot object detection in a cross-dataset setting from COCO to VOC to demonstrate the domain adaptation ability of our method,  where the domain of the novel model training dataset is different from the base model training dataset. 
We only report the performance of the novel classes since there are no base classes in VOC. As shown in Table~\ref{ifsod on cross dataset}, our method achieves much better result of $16.6\%$ AP compared to ONCE~\cite{perez2020incremental} with $2.4\%$ AP in the 5-shot setting. Furthermore, 
we can see that our method also significantly outperforms ONCE (Ours: $24.6\%$ vs. ONCE: $2.6\%$) in the 10-shot setting.

\begin{table}[!htbp]
\large
\centering
\resizebox{0.9\linewidth}{!}{
\begin{tabular}{p{1cm}<{\centering}|p{5cm}<{\centering}|p{1.5cm}<{\centering}p{1.5cm}<{\centering}}
\hline
\multirow{2}{*}{Shot} & \multirow{2}{*}{Method} & \multicolumn{2}{c}{Novel}     \\ \cline{3-4} 
                       &                         & \multicolumn{1}{p{2cm}<{\centering}|}{AP(\%)} & AP50(\%) \\ \hline
                      
\multirow{2}{*}{1}      &ONCE~\cite{perez2020incremental}                         & \multicolumn{1}{c}{-}   &-      \\  
                       &Ours                       & \multicolumn{1}{c}{4.1}   & 6.6     \\ \hline
                      
\multirow{2}{*}{5}      & ONCE~\cite{perez2020incremental}                        & \multicolumn{1}{c}{2.4}   &-      \\  
                       &  Ours                       & \multicolumn{1}{c}{\textbf{16.6}}   & \textbf{26.3}     \\ \hline
                      
\multirow{2}{*}{10}      &ONCE~\cite{perez2020incremental}                         & \multicolumn{1}{c}{2.6}   &-      \\  
                       & Ours                        & \multicolumn{1}{c}{\textbf{24.6}}   & \textbf{38.4}     \\ \hline
\end{tabular}}
\caption{\small Results of incremental few-shot object detection on VOC $test$ set.} 
\label{ifsod on cross dataset}
\end{table}

\subsection{Ablation Studies}

Table~\ref{ablation} shows the ablation studies to understand the effectiveness of each component in our framework. 
The first row is the result of the model initialized from the pre-trained base model, and then trained with the standard Deformable DETR loss on a few novel samples without using any component introduced in our framework.
We can see that this setting gives the lowest performance, which is evident on the importance of the components of our method. The subsequent settings in Row 2-8 include various combinations of the components from our framework. 
Our complete framework achieves the best performance of all classes.
These results further show the effectiveness of the two-stage fine-tuning, self-supervised learning and knowledge distillation strategies in our method.

We combine the two steps from the base model pre-training stage into a single step and train the model from scratch with the pseudo targets generated by the selective search algorithm and the targets from the abundant base class data. 
As shown in Row 1 of Table~\ref{ablation2}, we can see that the results of base and novel classes drop (Ours: 30.5\% vs One-stage: 28.3\% AP for base classes, Ours: 8.3\% vs One-stage: 7.1\% AP for novel classes) in one-step base model pre-training stage.
Furthermore, we report the effectiveness of keeping the class-agnostic modules of the model frozen in the incremental few-shot fine-tuning stage. As shown in Row 2 of Table~\ref{ablation2}, we can see that the results of base and novel classes drop significantly when unfreezing the class-agnostic CNN backbone, transformer and regression head 
in incremental few-shot learning stage. These results demonstrate that each component in our method has a critical role for incremental few-shot learning.

\section{Conclusion}
\label{conclusion}
In this paper, we present a novel incremental few-shot object detection framework: the Incremental-DETR for the more challenging and realistic scenario with no samples from the base classes and only a few samples from the novel classes in the training dataset.
To circumvent the challenges in DETR for incremental few-shot object detection, we identify that the projection layer and the classification layer of the DETR architecture are class-specific and other modules like the transformer are class-agnostic with empirically experimental studies. We then propose the use of two-stage fine-tuning strategy and self-supervised learning to retain the knowledge of the base classes and to learn a better generalizable projection layer to adapt to the novel classes.  
Furthermore, we utilize a knowledge distillation strategy to transfer knowledge from the base to the novel model using the few samples from the novel classes. 
Extensive experimental results on the benchmark datasets show the effectiveness of our proposed method.
We hope our work can provide good insights and inspire further researches in the relatively under-explored but important problem of incremental few-shot object detection.

\section{Acknowledgements}
\label{acknowledgement}
The first author is funded by a scholarship from the China Scholarship Council (CSC). This research/project is supported by the National Research Foundation Singapore and DSO National Laboratories under the AI Singapore Programme (AISG Award No: AISG2-RP-2020-016) and the Tier 2 grant MOE-T2EP20120-0011 from the Singapore Ministry of Education.


\end{document}